# Représentation formelle des concepts spatiaux dans la langue

Michel Aurnague*, Laure Vieu**, Andrée Borillo*

## Introduction

Dans ce chapitre, nous faisons l'hypothèse que l'étude systématique de la sémantique des marqueurs spatiaux de la langue permet de mettre en évidence certaines propriétés et concepts fondamentaux caractérisant les représentations conceptuelles de l'espace. Nous proposons un système formel rendant compte des propriétés révélées par les analyses linguistiques, et nous utilisons ces outils pour représenter le contenu sémantique de plusieurs relations spatiales du français. Le choix d'un système formel de représentation — en l'occurrence une théorie axiomatique dans le cadre de la logique des prédicats — répond à deux objectifs. D'une part, nous considérons qu'une sémantique doit prendre en compte la dimension déductive de la communication en langue naturelle et, de ce point de vue, la logique s'avère particulièrement adaptée à la formalisation du raisonnement. D'autre part, du fait de son caractère explicite, une représentation formelle fournit une base théorique adéquate à la réalisation d'un système de compréhension automatique de l'expression spatiale dans la langue.

L'utilisation de la langue comme base empirique permettant d'accéder à certains types de représentations conceptuelles et la prise en compte des aspects déductifs et inférentiels inscrivent bien ce travail dans une perspective cognitive.

La première partie du chapitre expose les principales propriétés révélées par l'analyse sémantique de l'expression de l'espace en français. Ces propriétés sont autant de contraintes que devront vérifier les représentations formelles élaborées. Nous présentons ensuite les diverses composantes du système de représentation formelle : les fondements théoriques d'une géométrie cognitive ou de sens commun (deuxième partie) sont d'abord esquissés, puis divers concepts fonctionnels et pragmatiques nécessaires au traitement de l'espace linguistique sont introduits (troisième et quatrième parties). Nous nous attachons à montrer comment ces concepts permettent de représenter le contenu sémantique de certaines prépositions du français (*sur, dans, devant*...) et à illustrer l'adéquation inférentielle de ces représentations.

## 1 Quelques propriétés de l'espace linguistique

### 1.1 Comment s'opère la localisation ?

Les langues naturelles sont généralement très riches en éléments lexicaux spécifiant la localisation des entités physiques dans l'espace, richesse non seulement en termes de nombre mais également pour la diversité des catégories lexicales concernées. En français, par exemple, toutes les catégories lexicales participent à l'expression des relations de type spatial:

• Certains noms et adjectifs fournissent la caractérisation des dimensions spatiales. Ils trouvent leur valeur au sein d'un système triaxial (vertical, frontal, latéral) fondé sur des données physiques du monde: gravité et verticalité, niveau du sol et horizontalité...

---

* Equipe de Recherches en Syntaxe et Sémantique (UMR 5610-CNRS), Maison de la Recherche, Université de Toulouse-Le Mirail, 5, allées Antonio Machado, 31058 Toulouse Cedex, email: aurnague@irit.fr, borilloa@irit.fr
** Institut de Recherche en Informatique de Toulouse (UMR 5505-CNRS), Université Paul Sabatier, 118, route de Narbonne, 31068 Toulouse Cedex, email: vieu@irit.fr

(*longueur, largeur, hauteur, profond, élevé, étroit*). D'autres intègrent des données perceptuelles et orientationnelles fournies par la situation de discours : position canonique du locuteur, définition de points de référence, choix d'évaluation égocentrique (*droite, gauche, avant, arrière, loin, proche*). Une catégorie particulière de noms et d'adjectifs (dits de localisation interne) ont pour fonction de préciser la référence spatiale des entités physiques [11]. La portion d'espace qu'occupe une entité déterminée peut être découpée en zones différenciées, sur la base des relations spatiales qui les rattachent au tout — et par conséquent qui les lient entre elles au sein de ce tout : *le haut, le bas; le bord, le centre; l'intérieur, l'extérieur; (zone) supérieure, (partie) centrale...* [4]. Ces noms et adjectifs fondés sur des traits de nature diverse (dimensionnels, morphologiques, fonctionnels) permettent de donner une plus grande précision à la description spatiale des entités :

> *Le haut de l'armoire est décoré ; la caisse a été endommagée dans sa partie supérieure*

• D'autres catégories lexicales, prépositions, verbes et adverbes, sont plus nettement spécialisés dans l'expression des relations spatiales, qui sont à la base même du principe de localisation. Ces relations ont un caractère statique lorsque les référents spatiaux, qu'ils soient immobiles ou en mouvement, sont considérés ponctuellement, dans la position qu'ils occupent à un instant donné :

> *La voiture est devant le camion ; le deuxième cheval est loin derrière le premier.*

Elles sont de caractère dynamique, s'il y a déplacement de l'un ou des deux référents, le déplacement induisant entre eux des modifications de leur relation spatiale :

> *La voiture passe devant le camion ; le deuxième cheval rattrape le premier.*

En français, très peu de prépositions de lieu sont spécialisées dans un emploi statique ou dynamique [12]. La plupart s'emploient aussi bien avec un verbe d'état —*être sur, se trouver à, se situer dans, être posé contre...*— qu'avec un verbe de mouvement —*aller à, monter sur, passer dans, buter contre...*— de sorte que la différence ne peut s'opérer qu'en présence du verbe avec lequel la préposition se construit :

> *Le chat est couché / se précipite **dans** le jardin.*
> *Les spectateurs sont assis /déambulent **autour** de la scène*

En revanche, pour les verbes, un partage assez tranché se fait entre ceux qui dénotent des relations spatiales de type statique, des verbes d'état en association avec des prépositions de lieu : *se trouver à, être placé dans, être posé sur, être appuyé contre...* et ceux de type dynamique qui eux se construisent avec ou sans préposition de lieu : *traverser, contourner, longer, entrer dans, passer sous, se glisser derrière...* Cependant, cette distinction entre statique et dynamique, dont on attendrait qu'elle se fonde sur des propriétés définies et stables, ne peut se faire parfois qu'au vu de l'énoncé auquel participe le verbe. Un même verbe prend une acception de type dynamique ou au contraire de type statique selon la nature et les propriétés du sujet de la phrase avec lequel il se construit, l'acception statique étant généralement liée à des facteurs d'ordre divers : attribution d'une valeur métaphorique, action de facteurs perceptuels, effet du traitement égocentrique de l'espace... Ainsi, on peut dire :

> *La souris/ la glycine court le long du mur; les promeneurs/ les rochers descendent jusqu'à la mer; le train/ le sentier s'enfonce dans la forêt*

La localisation à laquelle contribuent la plupart des prépositions et des verbes porte sur deux référents —entités physiques, lieux, portions d'espace plus ou moins délimitées— qui se définissent l'un par rapport à l'autre ( relation binaire dans laquelle cible et site trouvent chacun leur fonction) mais il existe également quelques cas de marqueurs relationnels mettant en jeu des référents dont le nombre est supérieur à deux :

• relation ternaire exprimée par des prépositions comme *entre, à l'intersection de, au confluent de, à équidistance de*…, par des verbes comme *séparer, relier, s'intercaler, s'interposer...* : le radiateur est entre la porte et l'armoire ; un mur sépare le jardin de la rue .

• relation affectant plus de trois entités exprimée également par *entre* mais aussi par *parmi, au milieu d*e (+pluriel) : *prendre place parmi les invités; avancer au milieu des voitures*. Il est clair qu'au-delà de deux référents, la relation spatiale que peuvent exprimer un verbe et/ou une préposition de lieu perd beaucoup de sa précision ; elle est tout au plus une vague indication de la localisation d'un référent , la cible, par rapport à des repères dont les propriétés spatiales n'ont rien de distinct (impossible de savoir s'il s'agit de points, de surfaces ou de volumes). Pour atteindre la précision que sont susceptibles d'apporter certaines relations binaires, il est nécessaire de cumuler les informations et donc d'avoir recours à des énoncés plus extensifs et plus complexes : *La voiture est entre le camion et l'autobus* peut signifier "*devant le camion et derrière l'autobus*" ou "*à droite du camion et à gauche de l'autobus*"...

### 1.1.1 L'imprécision du langage naturel : gêne ou avantage ?

Cette imprécision, tout à fait flagrante pour les relations autres que binaires, pose déjà un problème si l'on considère bon nombre de relations n'engageant que deux référents spatiaux. Bien sûr, on peut, si on le veut, être très précis et recourir à des coordonnées stables et à des valeurs numériques fixant des points sur ces coordonnées (données obtenues par la mesure des latitudes et des longitudes, par la mesure des degrés sur les axes d'orientation terrestre...) mais le plus souvent ces relevés très précis ne s'avèrent réellement nécessaires que pour des besoins de type scientifique ou technique (aéronautique, marine, météorologie...). Pour un usage plus courant, les expressions que nous pouvons forger à partir des ressources lexicales de la langue permettent de formuler des énoncés relativement clairs et informatifs, répondant la plupart du temps aux besoins et aux objectifs de la situation de communication telle que la conçoivent locuteur et destinataire.

Il est certain qu'un prédicat comme *être à côté de*, *être derrière*, *être devant* pris isolément, produit un sens très vague et très imprécis. Un énoncé comme *X est devant Y* ne dit pas à quelle distance X est de Y, ni avec quel décalage par rapport à son axe frontal. Cependant, si l'on connaît les entités physiques mises en jeu, la distance peut déjà trouver d'elle-même un certaine échelle de mesure. Si l'on dit :le verre est devant la bouteille , on peut imaginer qu'il s'agit d'une distance de quelques dizaines centimètres tandis que pour : la voiture est devant l'immeuble, on l'évaluera sans doute en mètres ou en dizaines de mètres. Pour ce dernier exemple, les choses peuvent encore se préciser, si au lieu du verbe *être* très général, on a recours à un verbe comme *être garé* ou *être rangé* : *la voiture est rangée devant l'immeuble*. Dans ce cas, notre connaissance du monde nous aide à affiner notre appréciation de *devant* en termes de position (calculée à partir des dimensions frontales du site) et de distance (que nous évaluerons en l'occurrence à quelques mètres, tout au plus).

La même stratégie de relativisation contextuelle est à l'oeuvre avec la plupart des prépositions de lieu de type relationnel:

*La mouche tourne au-dessus de l'assiette / l'avion tourne au-dessus de la ville;*
*La cuillère est à côté du couteau /Muret est à côté de Toulouse.*

Si dans des cas comme ceux-ci, l'imprécision peut être réduite grâce à des données de type pragmatique que nous savons intégrer et faire jouer pour l'interprétation, il n'en reste pas moins que les marqueurs linguistiques sont rarement capables de fournir l'information nécessaire pour déterminer sans erreur la position exacte d'un référent à localiser. Même lorsqu'on veut rendre compte de positions bien tranchées que l'on essaie de déterminer linguistiquement à l'aide d'adverbes de type *exactement, (tout) juste, carrément...*, il n'est pas sûr que l'énoncé soit totalement dénué d'ambiguïté pour celui qui le reçoit. Si l'on dit : *l'arrêt (de l'autobus) est juste devant la gare* , veut-on vraiment signifier que l'arrêt se trouve exactement sur l'axe frontal tracé à partir du point figurant le centre de la gare ? rien n'est moins sûr, et le destinataire du message aurait sans doute tort de le prendre au pied de la lettre.

Mais la communication linguistique dans ce qu'elle a de plus courant et de plus banalement fonctionnel, ne nécessite pas une si grande rigueur dans la précision des données. Dans la plupart des cas, les indications approximatives que nous produisons à travers les très nombreuses expressions langagières dont nous disposons suffisent pour guider la compréhension et aider à la construction d'une représentation de la situation d'ensemble. A la limite, des indications trop affinées et trop précises pourraient s'avérer inefficaces car elles chargeraient inutilement le message et risqueraient de ralentir et de gêner le processus de production et d'interprétation. Parmi les principes gricéens de coopération, on se souvient que figure la maxime de quantité : "*ne pas dire plus ou moins qu'il n'est nécessaire pour l'efficacité du message*". Il peut être contre-productif de fournir une trop grande spécification, qui peut se révéler inutile et être ressentie comme superflue par rapport à la situation particulière dans laquelle elle figure . Que penser par exemple de l'énoncé suivant :

> *Le vase est posé sur l'étagère, dans l'angle de droite, à 10 cm du bord du côté droit, à 15 cm du bord avant et à 5 cm du mur qui est derrière ?*

Cependant, si le besoin d'une information plus stricte est ressentie par le locuteur ou par le destinataire, des indications complémentaires peuvent s'ajouter pour spécifier certains traits de la configuration situationnelle.

> *A. Le titre en haut de la page*
> *B. Où ? à quel niveau ?*
> *A. A gauche, décalé de 1 cm par rapport à la marge et avec un espace de trois interlignes par rapport à la première ligne du texte.*

On notera que ces indications sont nécessairement d'ordre relationnel, quel que soit le degré de précision que l'on tente d'introduire : à 3 cm du haut de la page, à 3 cm du bord supérieur de la page et à 5 cm du bord latéral.

## 1.1.2 Définition des propriétés spatiales et notion de point de vue : le choix d'une granularité variable.

Il semble normal de vouloir définir les référents spatiaux à partir de leurs propriétés dimensionnelles de base, c'est-à-dire les définir comme des points, des lignes, des surfaces ou des volumes. Cependant, la vision que nous avons de ces référents, le type d'intérêt que nous leur portons lorsque nous en parlons fait que nous pouvons les considérer sous des angles différents et donc leur attribuer momentanément des propriétés dimensionnelles différentes. On peut dire du même objet : *cette boîte est carrée/ronde*, si l'on s'arrête plus particulièrement à un aspect saillant de sa forme ou *cette boîte contient 2 kg de miel* si l'on s'intéresse à sa contenance. De même, on peut parler de la mer ou du soleil comme d'une surface : *la mer est plate aujourd'hui, le disque solaire touche l'horizon* , ou encore, on peut décrire une colline ou une montagne selon la forme de sa ligne de contour : *des collines douces; un massif aux arêtes vives, une chaîne très découpée*.

A la limite, une même entité peut être vue successivement comme un point, une surface ou un volume selon la distance à laquelle elle est perçue, selon la situation dans laquelle elle s'inscrit, selon la fonction à laquelle on s'intéresse. S'agissant d'une ville, on peut la voir comme un point, et la figurer comme telle sur une carte ou dans notre représentation mentale lorsque nous imaginons une vue d'ensemble à une très petite échelle, mais on peut également en parler comme d'une surface : *la ville s'étend sur un rayon de 10 km ; la ville couvre une superficie de 100 km$^2$* , ou comme d'une entité à trois dimensions: *nous pénétrons dans la ville par le sud; un plan est nécessaire pour s'orienter à l'intérieur de la ville* ou bien comme des deux à la fois : *nous pénétrons dans une ville qui a l'étendue d'une véritable capitale*

Plus largement, on sait que tout ce qui est voie de communication (rue, chemin, route, chemin de fer) et cours d'eau (rivière, canal, fleuve) est tantôt traité comme une ligne : *les courbes du chemin; le tracé du canal, la rue est toute droite, le fleuve est sinueux , la route s'étire,* tantôt comme une surface ou éventuellement comme un volume : *le chemin est cabossé, la route est asphaltée et bien plane, le canal a un volume d'eau bien faible ; s'engager dans une rue; plonger dans la rivière.* Sans compter qu'une même phrase peut référer à plusieurs de ces dimensions : *une route asphaltée et bien plane, au tracé rectiligne...*

Tous ces exemples sont l'illustration d'un même phénomène que révèle le langage en ce qui concerne notre traitement de l'espace et des référents spatiaux : la vision et la représentation que nous en avons n'est pas établie une fois pour toutes, sur des propriétés immuables et s'excluant mutuellement. Bien au contraire, vision et représentation fluctuent, se modifient selon les variations dues à un certain nombre de facteurs qui entrent obligatoirement en ligne de compte : données perceptuelles, évidemment, mais également et surtout données liées à la situation de discours (thématique, argumentation, finalité...). Ces points de vue différents que nous introduisons dans nos représentations et qui changent apparemment les propriétés des référents spatiaux que nous traitons induisent sur ces référents ce qu'il est convenu d'appeler une granularité variable, que nous devons nécessairement intégrer dans la représentation et dans le calcul.

## 1.2 Au-delà de la géométrie

Il apparaît déjà dans cette analyse de l'expression de la localisation spatiale en français ainsi que dans nombre de travaux antérieurs sur la sémantique des marqueurs linguistiques de l'espace (plus particulièrement [32, 8]) que les notions géométriques présentes dans la langue ne permettent pas à elles seules de représenter la sémantique de l'espace. Le contexte d'énonciation est, comme nous l'avons vu, un facteur que l'on ne peut négliger.

En outre, de nombreux phénomènes liés aux propriétés dites "fonctionnelles" des entités doivent être pris en compte. En effet, la description des dimensions et des formes de certaines entités et des relations géométriques entre ces formes, même contextualisée, ne suffit pas à déterminer l'applicabilité d'une expression spatiale. Si, par exemple, la sémantique de la préposition *sur* était représentée sur la base de la seule relation de contact, il ne serait pas possible de distinguer les configurations spatiales décrites par les phrases suivantes :
> *L'affiche est sur le mur (\*contre le mur)*
> *La planche est contre le mur (\*sur le mur)*

De la même manière, si l'inclusion de la cible dans la fermeture convexe du site décrivait complètement la préposition *dans*, on ne pourrait expliquer la raison pour laquelle la phrase suivante ne peut être utilisée pour décrire la situation correspondant à la figure 1 (exemple inspiré de [19]) :
> *L'abeille est dans le vase*

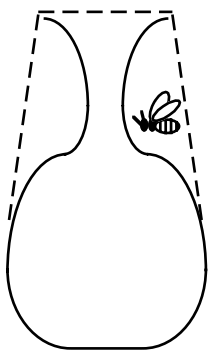

Figure 1

C'est parce qu'elles laissent de côté des concepts fonctionnels aussi fondamentaux que le support et la contenance que les définitions géométriques évoquées plus haut ne parviennent pas à rendre compte de la sémantique des prépositions *sur* et *dans*. De manière plus générale, la sémantique des marqueurs linguistiques de l'espace fait largement appel aux caractéristiques fonctionnelles des entités et des relations entre ces entités. Ces propriétés fonctionnelles peuvent appartenir au champ de la "physique naïve" (telle que définie dans [18]) comme dans le cas du support et de la contenance; elles peuvent également relever du domaine de l'orientation, les propriétés fonctionnelles liées à l'usage des entités jouant un rôle décisif dans l'interprétation d'expressions telles que *le haut de la bouteille, le devant de l'armoire* ou *l'aile gauche de la voiture*. Il existe enfin un certain nombre de propriétés fonctionnelles se rapportant à la structure interne des entités et spécifiant le caractère continu ou discret ou bien encore la structuration en parties ou composition. Ces dernières notions reposent sur une classification ontologique des entités qui est elle-même de nature essentiellement fonctionnelle.

Ces remarques, ainsi que d'autres observations du même type, nous ont amenés à adopter une approche en trois niveaux pour l'analyse et la représentation de la signification des expressions spatiales. Nous considérons tout d'abord un niveau **géométrique** représentant l'espace objectif décrit par le texte analysé ; ce niveau constitue la base du système. Un niveau **fonctionnel** prend ensuite en considération les propriétés des entités introduites par le texte et les relations non géométriques entre entités. Enfin, à un niveau **pragmatique**, sont introduites les conventions et les principes sous-jacents à une "bonne" communication [17] qui s'appuient pour l'essentiel sur des informations extérieures au texte lui-même telles que le contexte ou la connaissance du monde. Loin d'être indépendants, ces trois niveaux forment entre eux une structure hiérarchique : le second niveau introduit des informations fonctionnelles en se fondant sur des données géométriques (ainsi la contenance implique l'inclusion dans l'intérieur) et permet dès lors de représenter la sémantique "brute" des expressions spatiales. De son côté, le niveau pragmatique modifie les résultats obtenus au second niveau de manière à adapter cette sémantique à la situation "réelle".

Les éléments manipulés aux différents niveaux de notre système sont de nature distincte. Les phénomènes pragmatiques et fonctionnels prennent en compte les entités avec toutes leurs caractéristiques, leurs propriétés, leurs fonctions alors que les relations géométriques sont indépendantes de la couleur, la substance, l'usage... Il est ainsi possible de distinguer plusieurs entités décrivant la même portion d'espace-temps. Par exemple, *l'eau dans le verre* et *l'intérieur du verre* sont des entités bien différentes mais entretiennent les mêmes relations géométriques avec les autres entités, si bien qu'elles ne peuvent être distinguées au niveau géométrique. Les niveaux fonctionnel et pragmatique traitent donc directement des entités introduites par le texte alors que le niveau géométrique traite des référents spatiaux de ces entités, c'est-à-dire, les portions d'espace-temps qu'elles déterminent. On nommera les éléments du niveau géométrique à l'aide des termes d'"individus" ou de "corps", et ces éléments seront dénotés, à partir de la section 3 par des termes de la forme sref(x) où x est une entité du niveau fonctionnel, introduite par le texte.

## 2 Esquisse d'une géométrie cognitive

Dans la section précédente, nous avons pu voir que la langue décrit l'espace de façon relationnelle. Une représentation formelle de même type semble donc plus naturelle que l'emploi d'une géométrie de type cartésien où les individus sont situés au moyen de coordonnées. Mais ce n'est pas seulement plus naturel, c'est aussi plus efficace. En effet, les deux autres caractéristiques inhérentes aux expressions spatiales que nous avons décrites, leur imprécision et la variabilité de la "granularité" rendent l'usage direct d'un espace de points situés par coordonnées impossible. Il faudrait introduire des systèmes raffinés de gestion de l'imprécision et de l'incomplétude de l'information, ainsi que d'interprétation des points de vue dimensionnels. Une représentation directement relationnelle entre individus étendus est non seulement plus proche de l'expression langagière, mais satisfait pratiquement d'emblée les contraintes évoquées, la précision et la granularité s'ajustant à la richesse de la description en termes de relations et d'individus.

Nous allons voir dans cette section comment une telle représentation formelle peut être construite sur la base d'une nouvelle théorie de l'espace, la géométrie classique n'ayant pas développé cette voie. Même si la géométrie euclidienne est relationnelle, elle se fonde sur des éléments abstraits que sont les points, droites et plans, et ne peut traiter directement d'individus étendus. En suivant l'ordre de complexité attesté par des études en psychologie sur l'acquisition des concepts spatiaux par l'enfant [25], nous introduirons tout d'abord les relations du domaine de la méréologie et de la topologie, comme l'inclusion et le contact, puis celles relevant du domaine de la distance, et enfin celles concernant les principes d'orientation.

### 2.1 Méréologie et topologie

Un petit nombre de théories formalisant les concepts de la topologie sur des individus étendus ont été proposées par des logiciens [34, 30, 15], avec un actuel regain d'intérêt dans le domaine du raisonnement spatial qualitatif [28]. Elles sont toutes basées ou inspirées de la méréologie, la théorie de la relation d'inclusion exempte de la notion ensembliste d'élément [20, 29].

Notre théorie est une adaptation et extension de celle présentée dans [15]. Ce système méréo-topologique est construit à partir d'une primitive unique de "connexion", notée C. Deux individus sont connectés s'ils ont une partie en commun ou s'ils sont joints par une partie de leur surface. Le sens précis de cette relation est donné par les axiomes suivants :

La connexion est réflexive et symétrique :
A1 $\forall x\ C(x,x)$
A2 $\forall x\ \forall y\ (C(x,y) \rightarrow C(y,x))$

Deux individus sont (spatialement) égaux lorsqu'ils sont connectés aux mêmes individus (extensionnalité) :
A3 $\forall x\ \forall y\ (\forall z\ (C(z,x) \leftrightarrow C(z,y)) \rightarrow x =_s y)$

Plusieurs relations méréologiques peuvent alors être définies :
D1 $P(x,y) \equiv_{def} \forall z\ (C(z,x) \rightarrow C(z,y))$     "x est une partie de (est inclus dans) y"
D2 $PP(x,y) \equiv_{def} P(x,y) \land \neg P(y,x)$   "x est une partie propre de y"
D3 $O(x,y) \equiv_{def} \exists z\ (P(z,x) \land P(z,y))$     "x recouvre y"

La relation de jonction, également appelée "connexion externe", celle de partie tangentielle et celle de partie non tangentielle sont des relations qui ne relèvent pas de la méréologie. Leur définition, ainsi que celle des notions topologiques qui en sont dérivées, a été permise par le choix de la connexion comme primitive, au lieu de la relation d'inclusion P souvent choisie en méréologie classique :
D4 $EC(x,y) \equiv_{def} C(x,y) \land \neg O(x,y)$   "x est extérieurement connecté à y"
D5 $TP(x,y) \equiv_{def} P(x,y) \land \exists z\ (EC(z,x) \land EC(z,y))$     "x est une partie tangentielle de y"
D6 $NTP(x,y) \equiv_{def} P(x,y) \land \neg \exists z\ (EC(z,x) \land EC(z,y))$     "x est une partie non tangentielle de y"

Ces relations peuvent être schématisées par la figure 2:

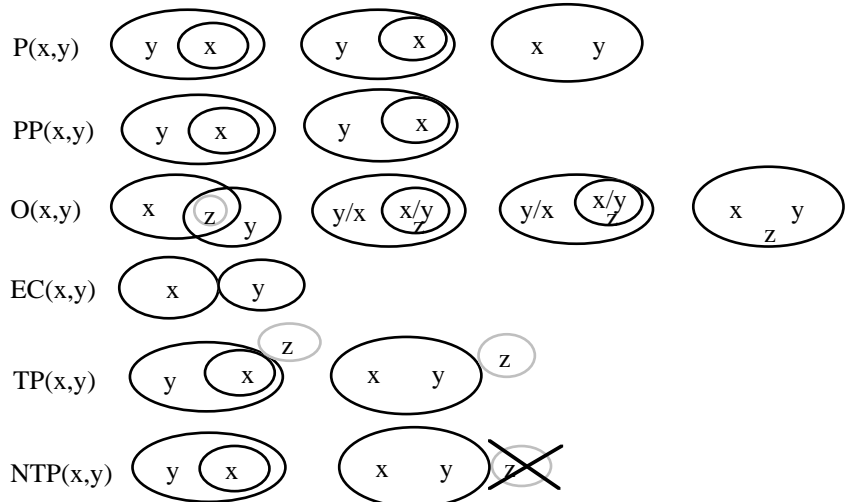

Figure 2

Ces définitions jointes aux axiomes donnent aux relations les propriétés inférentielles souhaitées : P, PP et NTP sont transitives, O et EC sont symétriques... Un grand nombre de propriétés sont aussi obtenues en combinant différentes relations ; on a par exemple : $\forall x\ \forall y\ \forall z\ ((NTP(x,y) \land EC(x,z)) \rightarrow O(y,z))$, c'est-à-dire que tout individu extérieurement connecté à une partie non tangentielle d'un autre individu recouvre aussi ce dernier.

La méréologie classique ainsi que la théorie de Clarke utilisent un opérateur général de fusion pour faire la "somme" d'une collection quelconque d'individus, principalement pour définir les opérateurs booléens d'union, intersection et complément. Il est en fait possible d'introduire axiomatiquement les opérateurs booléens directement, ce qui présente l'avantage important d'obtenir une théorie qui reste du premier ordre :

- 7 -

A4   $\forall x \forall y \exists z \forall u (C(u,z) \leftrightarrow (C(u,x) \vee C(u,y)))$

A4 et A3 impliquent, pour tout x et y, l'existence et l'unicité de leur somme, que l'on notera x+y.

A5   $\exists x \forall u\, C(u,x)$

A5 et A3 impliquent l'existence et l'unicité d'un individu universel, noté a*.

A6   $\forall x (\exists y\, \neg C(y,x) \rightarrow \exists z \forall u (C(u,z) \leftrightarrow \exists v (\neg C(v,x) \wedge C(v,u))))$

A6 et A3 impliquent, pour tout x≠a*, l'existence et l'unicité de son complément, noté -x.

A7   $\forall x \forall y (O(x,y) \rightarrow \exists z \forall u (C(u,z) \leftrightarrow \exists v (P(v,x) \wedge P(v,y) \wedge C(v,u))))$

A7 et A3 impliquent, pour tout x et y se recouvrant, l'existence et l'unicité de leur intersection (non vide), notée x•y.

Les concepts purement topologiques d'intérieur, d'ouvert et de fermeture sont définis par les axiomes ci-dessous[1] :

A8   $\forall x \exists y \forall u (C(u,y) \leftrightarrow \exists v (NTP(v,x) \wedge C(v,u)))$

A8 et A3 impliquent, pour tout x, l'existence et l'unicité de son intérieur, noté ix.

D7   $cx =_{def} -i(-x)$

Ainsi définie, la fermeture de x, cx, existe seulement si x≠a* (du fait de la présence de l'opérateur de complément). Cet opérateur c devient une fonction en ajoutant A9 :

A9   $c(a^*)=a^*$

D8   $OP(x) \equiv_{def} x =_s ix$              "x est ouvert"

D9   $CL(x) \equiv_{def} x =_s cx$              "x est fermé"

D'autres concepts topologiques comme la connexité d'un individu peuvent également être introduits :

D10  $Sp(x,y) \equiv_{def} \neg C(cx,cy)$              "x et y sont séparés"

D11  $Con(x) \equiv_{def} \neg \exists y \exists z (x =_s y+z \wedge Sp(y,z))$         "x est connexe"

Enfin, l'axiome suivant est nécessaire pour achever de spécifier la notion d'ouvert :

A10  $(OP(x) \wedge OP(y) \wedge O(x,y)) \rightarrow OP(x \bullet y)$

Il nous semble important de pouvoir établir une comparaison formelle entre les théories mathématiques connues et celle que nous proposons après les travaux de Clarke. Nous avons montré dans [3] que les modèles de cette théorie sont basés sur un espace topologique "classique", et en particulier que les concepts d'ouvert et de fermé correspondent bien à ceux de la topologie.

Si l'importance cognitive de ces concepts n'est peut-être pas claire à première vue, il semble qu'on puisse affirmer que les individus déterminés par des objets matériels et les "morceaux d'espace" (voir plus loin la section 3.1) qui les entourent sont de nature différente, notamment en ce qui concerne leurs surfaces, ce qui peut précisément s'analyser en termes de fermés et d'ouverts. De plus, ces concepts permettent de définir la notion de contact dans toute sa variété. La connexion externe peut être considérée comme une sorte de contact "fort" qui apparaît par exemple entre parties adjacentes d'une même entité connexe : la main / le poignet. On peut définir en outre le contact "intermédiaire", un contact sans connexion et toutefois total puisqu'on ne peut rien glisser entre les deux individus, comme entre un verre et son intérieur, ainsi que le contact "faible", un contact entre deux objets matériels non connectés, comme entre un livre et la table sur laquelle il est posé, probablement le prototype du contact.

D12  $ICont(x,y) \equiv_{def} \neg C(x,y) \wedge C(cx,cy)$         "x et y sont en contact intermédiaire"

D13  $WCont(x,y) \equiv_{def} \neg C(cx,cy) \wedge \forall z ((P(x,z) \wedge OP(z)) \rightarrow C(cz,y))$    "x et y sont en contact faible, x et y se touchent"

Ce dernier type de contact, bien que le plus courant, se trouve être le plus complexe car il est dépendant du niveau de granularité de la description. Une approche modale de la prise en compte de cette notion de granularité est développée dans [3].

Le prédicat Cont regroupe les trois types de contact :

D14  $Cont(x,y) \equiv_{def} EC(x,y) \vee ICont(x,y) \vee WCont(x,y)$    "x et y sont en contact"

---

[1] La notion mathématique d'intérieur n'a rien à voir avec la notion de sens commun d'intérieur fonctionnel que nous introduisons dans la section 3.4.

La dernière partie de la composante topologique de notre géométrie concerne la notion de frontière ou de limite, telle qu'elle s'exprime à travers l'usage de noms de localisation interne comme *surface*, *dessus*, *bord*, ou *pointe*. Il est facile de voir que les portions d'espace déterminées par ces limites sont bien des corps, c'est-à-dire des individus étendus du même type que les autres, et non des éléments de dimension inférieure. Les exemples suivants montrent que l'on conçoit les surfaces des objets comme ayant une certaine épaisseur, et les pointes une certaine surface :

> *la surface de la table est éraflée*
> *la pointe de ce crayon est émoussée*

Cette constatation ne signifie pas qu'il n'y a pas de différence entre ces individus. Chaque type de limite met en évidence la minimalité de l'individu selon un critère différent.

Nous définissons tout d'abord en D16 l'enveloppe x d'un individu y comme étant sa partie tangentielle minimale telle que tout individu extérieurement connecté à y est aussi extérieurement connecté à x et réciproquement. Cette enveloppe x est en fait la surface maximale de l'individu y.

D15  $Env'(x,y) \equiv_{def} TP(x,y) \land \forall z (EC(y,z)) \leftrightarrow EC(x,z))$
D16  $Env(x,y) \equiv_{def} Env'(x,y) \land \forall w (Env'(w,y) \rightarrow P(x,w))$   "x est l'enveloppe de y"

L'enveloppe d'une enveloppe est cette même enveloppe, ce qui évite le problème délicat d'avoir une primitive supplémentaire décidant a priori quels sont les individus "normaux", ou non-limites. On peut remarquer également que la condition de minimalité dans D16 implique que cette définition n'est opératoire (c'est-à-dire a une extension non vide) que dans un domaine atomique, par exemple, un domaine fini. Cette contrainte nous semble réaliste dans le contexte d'interprétation de textes qui est le nôtre.

A partir de la définition d'enveloppe, on peut ensuite caractériser les limites de premier type d'une entité comme étant les parties tangentielles de son enveloppe :

D17  $Lim1(x,y) \equiv_{def} \exists z (Env(z,y) \land TP(x,z))$      "x est limite 1 de y"

On peut réitérer le processus. Les limites de second type seront des parties d'un contour, lui-même introduit comme "frontière" maximale d'une partie de l'enveloppe par rapport au reste de l'enveloppe. Enfin, les limites de troisième type seront les extrémités d'une limite 2.

D18  $Contour'(x,y) \equiv_{def} \exists w,w' (Env(w,w') \land TP(y,w) \land \forall z (P(z,w) \rightarrow (EC(z,y) \leftrightarrow EC(z,x)))$
D19  $Contour(x,y) \equiv_{def} Contour'(x,y) \land \forall w (Contour'(w,y) \rightarrow P(x,w))$   "x est le contour de y"
D20  $Lim2(x,y) \equiv_{def} \exists w,w' (Contour(w,w') \land Lim1(w',y) \land TP(x,w))$   "x est limite 2 de y"
D21  $Ends'(x,y) \equiv_{def} \exists w,w' (Contour(w,w') \land TP(y,w) \land \forall z (P(z,w) \rightarrow (EC(z,y) \leftrightarrow EC(z,x)))$
D22  $Ends(x,y) \equiv_{def} Ends'(x,y) \land \forall w (Ends'(w,y) \rightarrow P(x,w))$      "x est la fusion des bouts de y"
D23  $Lim3(x,y) \equiv_{def} \exists w,w' (Ends(w,w') \land Lim2(w',y) \land TP(x,w))$   "x est une limite 3 de y"

Les frontières décrites par des expressions du type *la surface de la table*, *le bord de la table*, ou *le coin de la table* désignent en fait des individus ayant des propriétés supplémentaires par rapport à Lim1, Lim2 et Lim3 :

D24  $Surface(x,y) \equiv_{def} Con(x) \land Lim1(x,y) \land \neg Lim2(x,y)$      "x est une surface de y"
D25  $Line(x,y) \equiv_{def} Con(x) \land Lim2(x,y) \land \neg Lim3(x,y)$      "x est une ligne de y"
D26  $Point(x,y) \equiv_{def} Con(x) \land Lim3(x,y)$            "x est un point de y"

## 2.2 Distance

La distance est définie en mathématiques par une fonction qui, à deux points, ou par extension à deux ensembles de points, associe un nombre réel positif. Cette notion semble donc numérique par essence. Toutefois, comme pour bien d'autres concepts qui ont été modélisés par l'arithmétique, la notion sous-jacente essentielle est en fait un ordre, qui peut être modélisé symboliquement. Cette notion permet de traiter directement les différentes comparaisons de distance qui apparaissent dans la langue dans les expressions *plus près*, *plus loin*, *plus grand*, *plus petit* ou même dans des adjectifs de forme comme *carré* ou *rond*.[2]

Suivant ici la même démarche que précédemment, nous introduisons donc maintenant une nouvelle relation ternaire primitive entre individus, Closer(x,y,z), qui se lit "x est plus près de y que de z". Une relation similaire a été introduite dans [31], entre triplets de points. Comme nous allons le voir, une relation entre individus est plus complexe car elle interagit avec la méréo-topologie.

Closer(x,y,z) établit implicitement un ordre entre les couples d'individus (x,y) et (x,z). Cet ordre est strict (A11), ce qui permet de définir la relation d'équidistance par D27.

A11  Closer(x,y,z) → ¬Closer(x,z,y)
D27  Equidist(x,y,z) ≡$_{def}$ ¬Closer(x,y,z) ∧ ¬Closer(x,z,y) "x est à égale distance de y et de z"

L'ordre implicite est un ordre total (A12), qui est bien entendu transitif (A13 et A14).
A12  Closer(x,y,z) → (Closer(x,y,t) ∨ Closer(x,t,z))
A13  (Closer(x,y,z) ∧ ¬Closer(z,y,x)) → Closer(y,x,z)
A14  (Closer(x,y,z) ∧ ¬Closer(x,t,z)) → Closer(x,y,t)

La topologie induit des contraintes supplémentaires sur la notion de distance minimale:
A15  C(x,y) → ¬Closer(x,z,y)
A16  (C(x,y) ∧ ¬C(x,z)) → Closer(x,y,z)
A17  (WCont(x,y) ∧ ¬C(x,z)) → ¬Closer(x,z,y)
A18  (WCont(x,y) ∧ ¬WCont(x,z) ∧ ¬C(x,z)) → Closer(x,y,z)

Enfin, l'ordre de la distance est lié à celui de l'inclusion :
A19  P(x,y) → ¬Closer(z,x,y)

Un certain nombre de propriétés désirables peuvent être démontrées. Par exemple, la transitivité de la relation d'équidistance sous ses deux formes :
(Equidist(x,y,z) ∧ Equidist(x,z,t)) → Equidist(x,y,t)
(Equidist(x,y,z) ∧ Equidist(z,x,y)) → Equidist(y,x,z)

Ou encore, le fait que rien n'est plus près de soi que soi-même :
¬C(x,y) → Closer(x,x,y)

Par contre, pour exprimer l'inégalité triangulaire, propriété bien connue de la distance, il nous manque la notion d'alignement. Cette dernière est introduite dans le paragraphe suivant traitant de l'orientation.

## 2.3 Orientation et géométrie projective

Pour pouvoir traiter des relations d'orientation, nous complétons notre ontologie en introduisant la notion de direction. Ces nouvelles entités de notre langage formel seront notées Di (nous employons donc un langage du premier ordre typé). On peut cerner intuitivement la notion de direction en supposant que ces variables désignent des droites vectorielles orientées.

Nous introduisons une relation primitive entre directions Kd(D1,D2,D3) indiquant que "D1 est plus proche de D2 que de D3" (en termes de valeurs angulaires). Cette relation est, elle aussi, similaire à la primitive K dénotant la distance relative entre points axiomatisée dans [31], c'est pourquoi nous la notons Kd. Elle est irréflexive et transitive (et donc asymétrique) :
A20  ¬Kd(D1,D2,D2)
A21  (Kd(D1,D2,D3) ∧ Kd(D1,D3,D4)) → Kd(D1,D2,D4)

Comme dans le cas de la relation Closer, un second type de transitivité doit être établi :
A22  (Kd(D1,D2,D3) ∧ Kd(D3,D1,D2)) → Kd(D2,D1,D3)

---

[2]Les expressions de la distance en langage naturel font parfois usage de valeurs numériques, comme dans *Muret est à 20 km de Toulouse*. Nous pensons que tenter de traiter l'aspect numérique sans considérer l'aspect imprécis de ces expressions serait erroné puisque cette imprécision affecte l'opération numérique de base qu'est l'addition des distances. Comme le traitement de ce genre d'imprécision du langage naturel va bien au-delà du seul problème de la distance, nous gardons ce problème pour de futurs travaux. Cependant, ce n'est pas trop s'avancer que de penser que la modélisation qualitative de la distance que nous proposons ici sera alors utile.

La primitive Kd permet de caractériser les notions de directions opposées et de directions orthogonales. L'opposée d'une direction est la direction la plus éloignée de cette dernière alors qu'une direction orthogonale à une direction donnée est située à égale distance de cette direction et de son opposée :

D28  $-(D1,D2) \equiv_{def} D3 \neq D2 \rightarrow Kd(D1,D3,D2)$
D29  $Ortho(D1) =_{def} \{D2: -D1=D3 \land \neg Kd(D2,D1,D3) \land \neg Kd(D2,D3,D1)\}$

Un axiome supplémentaire assure l'existence de l'opposée d'une direction :
A23  $\forall D1 \exists D2 (\forall D3\ D3 \neq D2 \rightarrow Kd(D1,D3,D2))$.

Comme pour les axiomes A4-8, l'axiome A23 et le fait que l'opposée d'une direction est unique (ce qui peut être démontré en utilisant A23 et l'asymétrie de Kd), permettent d'introduire dans les notations un nouvel opérateur - sur les directions, -D dénotant la direction opposée à la direction D.

On peut alors définir la médiane de deux directions ainsi qu'une opération de somme ou de composition de directions. La somme de deux directions est le sous-ensemble de l'ensemble des médianes constitué par les directions qui sont les plus proches des deux directions considérées (pour des directions non opposées cet ensemble est un singleton alors que dans le cas de directions opposées cet ensemble comprend deux éléments en deux dimensions et définit un plan en trois dimensions). Nous introduisons ci-dessous les définitions caractérisant les médianes et les sommes ainsi qu'un axiome de linéarité :

D30  $Med(D1,D2) =_{def} \{D3: (D1=D2 \land D3=D1) \lor (D1 \neq D2 \land \neg Kd(D3,D1,D2) \land \neg Kd(D3,D2,D1))\}$
D31  $D3 \in Sum(D1,D2) \leftrightarrow (D3 \in Med(D1,D2) \land \forall D4\ (D4 \in Med(D1,D2) \rightarrow \neg Kd(D1,D4,D3)))$
A24  $(D1 \neq D2 \land D1 \neq D3 \land D2 \neq D3) \rightarrow (Kd(D1,D2,D3) \lor Kd(D1,D3,D2) \lor D1 \in Med(D2,D3))$

Deux axiomes expriment le caractère circulaire ou réflexif des directions :
A25  $Kd(D1,D2,D3) \leftrightarrow Kd(D1,-D3,-D2)$
A26  $Kd(D1,D2,D3) \leftrightarrow Kd(-D1,-D2,-D3)$

Enfin, un axiome établissant la transitivité entre médianes est introduit et la relation entre une direction D et deux directions D2 et D3 est exprimée sur la base de la somme de ces directions :
A27  $(D \in Med(D1,D2) \land D \in Med(D2,D3) \land D1 \neq D3) \rightarrow D \in Med(D1,D3)$
A28  $(Kd(D,D2,D3) \land D1 \in Sum(D2,D3)) \rightarrow (Kd(D3,D1,D) \land Kd(-D2,-D1,D))$

La théorie basée sur la primitive Kd comporte d'autres définitions et axiomes (directions coplanaires, extensionnalité...) et permet d'établir de nombreux théorèmes [5].

La formalisation des phénomènes orientationnels dans la langue nécessite également l'utilisation, au niveau géométrique, d'un ensemble de treize relations constituant une extension des relations d'Allen [1][3]. Chaque relation du type Rel(x,y,D) dénote la configuration dans laquelle se trouvent les intervalles maximaux définis par les individus x et y dans la direction D. Hormis les axiomes classiques associés aux relations d'Allen, nous introduisons ici un postulat établissant que pour toute paire d'individus connectés x et y et toute direction D, l'une des relations m, o, s, d, f ou = est vérifiée :
A29  $C(x,y) \rightarrow mosdf=m_io_is_id_if_i(x,y,D)$[4]

---

[3]Cette axiomatique, basée sur 13 relations mutuellement exclusives, a été proposée dans le but d'effectuer des calculs sur les intervalles temporels. <(x,y) dénote que x précède (complètement) y, m(x,y) que x (précède et) rencontre y, o(x,y) que x (précède et) chevauche y, s(x,y) que x débute y, f(x,y) que x termine y, et d(x,y) que x est inclus dans y (sans débuter ni terminer y). >, mi, oi, si, fi et di sont les relations inverses. x=y dénote l'égalité de x et y.

[4]Sur la base de ce postulat et en utilisant la définition de l'inclusion (P) ainsi que plusieurs théorèmes associés aux relations de Allen, il est possible par exemple de déduire que : $P(x,y) \rightarrow sfd=(x,y,D)$

Nous posons alors que y constitue une extrémité de x dans la direction D si y est une limite de x et si, de plus, tout individu inclus dans x (et non inclus dans y) précède ou rencontre y dans cette direction D :

D32  $Ext(y,x,D) \equiv_{def} Lim1(y,x) \land \forall v\ ((P(v,x) \land \neg P(v,y)) \to <m(v,y,D))$

Soulignons que, dans certaines situations, plusieurs directions peuvent vérifier cette relation pour deux individus x et y donnés. Généralement, ceci se produit lorsqu'une tangente à la surface ne peut être définie au point considéré (par exemple lorsque l'on se trouve en présence d'un sommet y d'un triangle x).

Si nous souhaitons qu'une seule direction soit sélectionnée, il est alors nécessaire d'introduire des contraintes supplémentaires dans la définition. Ceci nous amène à définir une relation "Exts" indiquant que y constitue une extrémité de x dans la direction D et z une extrémité (d'une partie u de x) dans la direction opposée :

D33  $Exts(y,z,x,D) \equiv_{def} Ext(y,x,D) \land \exists u\ (P(u,x) \land P(y,u) \land Ext(z,u,-D)$
     $\land\ Salient(z,x) \land (\neg \exists v\ Point(z,v) \lor \neg \exists v\ Point(y,v)))$

Dans cette définition, le prédicat "Salient" rend compte des processus visuels et cognitifs qui conduisent à sélectionner un individu géométriquement saillant z dans l'individu x. Le reste de la définition garantit que l'individu z constitue une extrémité dans la direction -D et que l'une des extrémités considérées n'est pas ponctuelle.

## 2.4 Vers une géométrie basée exclusivement sur les individus

L'approche formelle développée ici permet de saisir les notions topologiques et celle de distance qualitative sur la base d'une seule catégorie d'éléments primitifs, les individus. Ceux-ci correspondent aux morceaux d'espace (tridimensionnels ou étendus) que déterminent les entités de notre monde, parmi lesquels les objets. Ce choix nous semble non seulement justifié d'un point de vue cognitif mais également raisonnable d'un point de vue ontologique. En effet, contrairement à Aristote ou Kant, et en accord avec Leibniz, nous pensons que l'espace linguistique et cognitif n'est pas une structure abstraite donnée a priori mais qu'il est construit relationnellement à partir des entités qui nous entourent. Il est donc naturel que ces entités constituent le substrat de notre théorie de l'espace.

Toutefois la partie orientationnelle de cette théorie fait appel à un deuxième type d'éléments primitifs à savoir les directions. Afin de préserver la minimalité de l'ontologie nous envisageons d'étudier dans quelle mesure les directions pourraient être définies à partir des individus. Il semble possible par exemple d'introduire une relation d'alignement entre individus, une direction correspondant alors à un triplet d'individus alignés. Une autre possibilité serait la définition des directions comme étant essentiellement dynamiques et résultant du mouvement des individus. Un tel choix nécessite cependant la prise en compte des observations et des théories que proposent la psychologie cognitive et la psycholinguistique. Il serait donc important de déterminer quel(s) point(s) de vue, "abstrait" (directions vectorielles primitives, comme la gravité) ou "matériel statique" (direction donnée par l'alignement d'individus, par exemple établi par la vision d'objets en occultant d'autres), ou enfin "matériel dynamique" (linéarité du mouvement perçue grâce à la persistance rétinienne) sous-tend(ent) notre représentation mentale de l'orientation dans l'espace, si tant est qu'il y en ait un de primitif.

Ces considérations mises à part, ce travail n'est toutefois qu'une étape dans l'entreprise d'élaboration d'une géométrie cognitive. Il demande, en particulier, que des liens inférentiels soient établis entre ses trois parties, à savoir la topologie, la distance et l'orientation. Ces liens, c'est-à-dire très certainement des axiomes supplémentaires, permettront l'expression de propriétés comme l'inégalité triangulaire, mentionnée plus haut.

## 3 Concepts fonctionnels et relations spatiales

Comme cela a pu être mis en évidence dans la section 1.2, le fonctionnement des expressions spatiales fait largement appel aux propriétés fonctionnelles des entités et des relations entre ces entités. Nous analysons dans la suite un certain nombre de ces concepts fonctionnels. Il s'agit de la structure interne des entités, des notions d'orientation ainsi que des notions de support et de contenance.

## 3.1 Structure des entités et relations de partie à tout

Un certain nombre d'expressions spatiales comme les noms de localisation interne réfèrent à des parties d'entités, et donc à une structure interne de ces entités. On observe également que les prépositions dites topologiques comme *dans* et *sur* sont parfois utilisées pour décrire des relations de partie à tout, comme dans *les pépins sont dans la pomme* ou *les touches sont sur le clavier*. Plus généralement, la prise en compte de propriétés fonctionnelles comme celles donnant lieu à une orientation intrinsèque, repose sur l'analyse différenciée du rôle des parties dans le tout. La relation d'inclusion P relie parties et touts spatialement mais non fonctionnellement. Elle ne permet pas de distinguer entre différents types de relations de partie à tout (par exemple distinguer entre la relation entre une page et un livre et la relation entre la préface et le livre). La langue fait pourtant appel à des relations structurelles variées qui se distinguent notamment par leur comportement inférentiel [35].

Cette section propose une analyse des relations de partie à tout qui sous-tendent la notion de structure interne des entités, sans laquelle l'étude des rôles fonctionnels mis en jeu par les expressions spatiales ne saurait être complète. Nous y précisons également l'ontologie des entités apparaissant dans les expressions spatiales non métaphoriques considérées. En effet, les relations structurelles sont souvent manifestées par un choix linguistique qui correspond à un point de vue ontologique particulier sur les entités. Ainsi, le terme de masse *du riz* (ou *un tas de riz*) met en évidence une structure interne continue, alors que le terme pluriel *des grains de riz* introduit une structure de collection[5]. Un première tâche à réaliser est donc la modélisation de la notion de structure présente au niveau du syntagme nominal.

### 3.1.1 Structure plurielle

Les syntagmes nominaux pluriels (*Jean et Marie*, *les arbres*) réfèrent à des collections, ainsi que nombre de syntagmes nominaux singuliers (*le couple Dupont*, *la forêt*). Des études poussées sur les pluriels et la notion de collection ont été effectuées en sémantique formelle. Nous reprenons ici la structure de treillis introduite dans [21] que nous modifions afin, entre autres, de prendre en compte les remarques de [10].

Dans cette structure dite "plurielle", les atomes représentent les entités dénotées par un syntagme nominal singulier, qu'elles soient des collections ou non, et les constituants non-atomiques représentent les entités dénotées par un syntagme pluriel. La relation d'ordre du treillis lie donc les collections plurielles à leurs éléments (constituants inférieurs atomiques) et à leurs sous-collections (constituants inférieurs non-atomiques). Par contre, une collection singulière comme *la forêt* n'est pas liée directement à ses éléments dans le treillis, puisque c'est un atome. Elle l'est cependant indirectement, car il existe toujours une collection plurielle correspondante (dans cet exemple, *les arbres*), liée à ses atomes dans le treillis.

La relation primitive utilisée est un ordre partiel non strict, noté ≤. Les axiomes et définitions suivants sont nécessaires pour caractériser cet ordre :

A30 $(x \leq y \wedge y \leq z) \rightarrow x \leq z$
A31 $(x \leq y \wedge y \leq x) \leftrightarrow x = y$
D34 $At(x) \equiv_{def} \forall y \, (y \leq x \rightarrow y = x)$                    "x est atomique"
A32 $\forall x \, \forall y \, \exists z \, \forall u \, (u \leq z \leftrightarrow \exists v \, (v \leq u \rightarrow \exists w \, (w \leq v \wedge (w \leq x \vee w \leq y))))$     "z, noté $x \cup y$, est la somme de x et de y"
A33 $\forall x \, \forall y \, (\exists v \, (v \leq x \wedge v \leq y) \rightarrow \exists z \, \forall u \, (u \leq z \leftrightarrow (u \leq x \wedge u \leq y)))$        "z, noté $x \cap y$, est l'intersection de x et de y"
A34 $x \leq y \rightarrow P(sref(x), sref(y))$

Modéliser deux sortes de collections (plurielles et singulières) peut sembler arbitrairement compliqué. En fait, cela autorise la distinction entre plusieurs collections

---

[5]Bien qu'ils puissent décrire une même réalité physique, nous considèrerons dans ce cas, comme dans nombre de cas similaires étudiés dans la suite, que ces deux termes désignent bien deux entités différentes dans notre conceptualisation du monde liée à la langue, ce qui est bien l'objet de notre modélisation.

segmentant un même matériau, distinction qui apparaît dans la langue par exemple entre *les cartes* et *les jeux de cartes*, entités qui n'ont pas les mêmes éléments, même lorsqu'elles désignent une même réalité physique et ont donc le même référent spatial (exemple tiré de [21]). Cependant cette remarque pointe aussi sur le fait que ce lien spatial entre collection singulière et collection plurielle n'est pas univoque : dans ce même exemple, la collection singulière *le paquet de cartes* a le même référent spatial que les deux collections plurielles *les cartes* et *les jeux de cartes*. Inversement, certaines entités atomiques sont liées spatialement à des collections plurielles alors qu'elles ne sont pas des collections singulières. Par exemple, les entités décrites par les termes de masse *du riz* ou *le bol de riz* ne sont pas des collections singulières mais ont le même référent spatial que *les grains de riz*. Le lien spatial ne suffisant donc pas à établir les bonnes correspondances entre entités singulières et entités plurielles, nous introduisons une nouvelle relation primitive notée Is-coll(x,y), qui se lit "x est la collection des y". Elle vérifie les axiomes suivants :

A35 $\text{Is-coll}(x,y) \rightarrow (\text{At}(x) \wedge \neg \text{At}(y))$
A36 $\text{Is-coll}(x,y) \rightarrow \text{sref}(x) =_s \text{sref}(y)$
A37 $(\text{Is-coll}(x,y) \wedge \text{Is-coll}(x,z)) \rightarrow y=z$

Il est utile d'ajouter la définition suivante pour les collections :
D35 $\text{Coll}(x) \equiv_{\text{def}} \neg \text{At}(x) \vee \exists y\ \text{Is-coll}(x,y)$

### 3.1.2 Structure massique

En ce qui concerne les syntagmes nominaux qui sont des termes de masse, notre formalisation reprend surtout, parmi l'abondante littérature, les travaux de Parsons [23, 24].

Si pour traiter des pluriels, nous séparions implicitement les entité en deux classes, les collections et les non-collections (ou entités simples), nous devons ici introduire des distinctions ontologiques parmi les entités simples. Les termes de masse sont formés d'un déterminant partitif ou de mesure (du, de la, un peu de, un verre de...) et d'un nom de substance (eau, neige, sable, mobilier...). Afin d'expliquer correctement à la fois le comportement linguistique des termes de masse et celui des emplois nominaux génériques des substances (comme dans *l'eau est un liquide, l'oignon a un goût prononcé*) il est nécessaire de considérer que les substances sont toujours des entités simples particulières. De même, il convient de distinguer parmi les entités simples les quantités de substance, ou morceaux de matière, que les termes de masse permettent de désigner. Ces dernières ont la particularité de ne pas être comptables : *de l'eau* plus *de l'eau* est encore *de l'eau*. Cette propriété est souvent décrite sous le nom de "référence cumulative" [27].

Nous introduisons donc trois nouveaux prédicats : Subst(x) qui caractérise les substances, Mat(x) qui caractérise les morceaux de matière, et Q(x,y), relation introduite dans [23], que l'on peut paraphraser en "x est une quantité de y". Les axiomes suivants précisent leurs rapports entre eux et avec la structure spatiale :

A38 $Q(x,y) \rightarrow (\text{Mat}(x) \wedge \text{Subst}(y) \wedge P(\text{sref}(x),\text{sref}(y)))$
A39 $\text{Mat}(x) \rightarrow \exists y\ (Q(x,y) \wedge \forall z\ (Q(x,z) \rightarrow y=z))$
A40 $(Q(x,y) \wedge Q(z,y)) \rightarrow \exists t\ (Q(t,y) \wedge \text{sref}(t)=_s\text{sref}(x)+\text{sref}(z))$[6]
A41 $(\neg \text{Coll}(x) \wedge \neg \text{Coll}(y) \wedge \text{sref}(x)=_s\text{sref}(y) \wedge \exists z(Q(x,z) \wedge Q(y,z))) \rightarrow x=y$

### 3.1.3 Classification des entités

Nous venons de voir que les structures plurielles et massiques introduisent des notions qui reposent sur la distinction de différentes classes d'entités. Nous allons maintenant considérer dans son ensemble la classification que nous utilisons ici, autrement dit, l'ontologie de notre système formel.

---

[6]Cet axiome formalise la propriété de référence cumulative. La propriété de référence partitive (toute entité spatialement incluse dans une quantité de substance est aussi une quantité de cette substance), a souvent été décrite à propos des termes de masse. Nous ne la retenons pas ici, car les contre-exemples sont nombreux (*un pied de chaise* n'est pas *du mobilier*, *un atome d'hydrogène* n'est pas *de l'eau*) et cela rentrerait en contradiction avec le fait qu'on peut distinguer plusieurs entités pour un même référent spatial.

Cette classification a deux dimensions : la première divise les entités selon leur nombre et la seconde les divise selon leur essence. En ce qui concerne le nombre, nous avons vu que les entités peuvent être simples ou collectives, et que les entités collectives se répartissent en collections singulières et collections plurielles. En ce qui concerne la nature essentielle des entités, nous avons déjà rencontré au moins deux classes, les substances et les morceaux de matière. Pour analyser la sémantique des prépositions spatiales (on le verra en particulier pour *dans*) il est en fait nécessaire de considérer au total cinq classes : les objets (*Marie*, *une forêt*, *le bord de la table*), les morceaux de matière (*un verre d'eau*, *du mobilier*, *le bois de la chaise*), les substances (*la neige*), les lieux (*Toulouse*, *mon jardin*) et les morceaux d'espace (*l'intérieur d'une boîte*, *un trou dans le gruyère*, *une grotte*). Les objets (Obj) sont des entités comptables, matérielles, non génériques, en général mobiles. Les lieux (Loc) sont des entités fixes les unes par rapport aux autres, et pour lesquelles nous considérerons ici, mais c'est une simplification (cf. [6]), qu'elles sont co-extensionnelles avec une portion de la surface terrestre. Les morceaux d'espace (Sp-port) sont les seules entités immatérielles. Elles sont cependant toujours dépendantes, souvent de manière fonctionnelle, d'une ou plusieurs autres entités qui sont matérielles [14].

A42  $Sp\text{-}port(x) \rightarrow \exists y\ ((Obj(y) \vee Mat(y) \vee Loc(y)) \wedge Depend(x,y))$

Si les référents spatiaux des objets, morceaux de matière, substances et lieux, sont déterminés directement par leur extension matérielle, les référents spatiaux des morceaux d'espace sont déterminés indirectement par des fonctions géométriques sur les référents spatiaux des entités dont ils dépendent.

Nous avons pu noter que la classification ne dépend pas d'une réalité objective du monde, mais de la façon dont nous décrivons cette réalité dans la langue. La classification n'est cependant pas réalisée au niveau du lexique, un même lexème pouvant, selon l'usage, désigner des entités de nature différente. *Pomme* peut désigner un objet (*une pomme*), un morceau de matière (*de la pomme*) ou une substance (*la pomme* dans *la pomme et le hareng s'accordent bien*). De même, c'est le contexte qui permettra de déterminer si *la forêt* désigne l'objet collection d'arbres ou le lieu où cette collection pousse.

Même si les deux dimensions de la classification sont orthogonales, nous pensons nécessaire d'ajouter une contrainte entre les deux. Nous faisons l'hypothèse que les cinq classes "essentielles" sont séparées et épuisent bien les entités simples de notre domaine d'étude[7]. Ceci revient à supposer qu'il n'existe pas de lexème décrivant des collections hétéroclites. Nous ajoutons donc l'axiome suivant, où ⊕ dénote le ou exclusif :

A43  $At(x) \rightarrow (Obj(x) \oplus Mat(x) \oplus Subst(x) \oplus Loc(x) \oplus Sp\text{-}port(x))$

### 3.1.4 Méronomies

Grâce aux outils formels que nous venons d'introduire, il est possible de donner une définition pour les différentes relations de partie à tout, encore appelées méronomies, que la langue permet d'exprimer. Le classement des méronomies que nous introduisons ici est inspiré de [35] et est amplement motivé dans [33].

Nous avons déjà implicitement mentionné les deux méronomies "élément / collection" (*un arbre de la forêt*) et "sous-collection / collection" (*le conseil de sécurité de l'ONU*) que la structure plurielle permet de formaliser immédiatement.

La structure massique permet également de formaliser assez directement deux autres méronomies : "portion / tout" (*ceci est une part de gâteau*) et "substance / tout" (*il y a du sucre dans ce gâteau*). Pour la première relation, la partie et le tout sont deux quantités de la même substance, alors que dans le cas de la méronomie "substance / tout", deux substances, l'une pour la partie, l'autre pour le tout, sont mises en relation.

Deux autres relations méronomiques sont employées en français, mais cette fois, les structures plurielle et massique n'aident pas à leur formalisation. Ces méronomies relient des entités simples qui ne sont ni des morceaux de matière, ni des substances.

---

[7]Dans un travail ayant une portée spatio-temporelle, d'autres classes seraient introduites pour les éventualités (événements et états) et les temps (*lundi*, *cette année*...). Rappelons que nous ne considérons ici aucune entité abstraite.

La première, "composant / assemblage", est peut-être celle qui peut être considérée comme le prototype des relations de partie à tout (*le pied de la chaise*, *le moteur de la voiture*, *la main de mon bras droit*...). Elle fait surtout appel au fait que la partie remplit une fonction par rapport au tout [16], cette fonction étant évoquée par les termes employés pour désigner la partie et le tout. Nous n'analyserons pas ici la relation de fonctionnalité. Sa complexité, due au fait que toutes sortes de fonctions (support, production d'énergie, préhension...) peuvent être en jeu, est évidente.

La dernière de nos méronomies, "morceau / tout", est en contraste avec "composant / assemblage" justement en ce qui concerne l'absence de fonction évoquée. Si un composant a généralement une forme et une position déterminée par sa fonction, un morceau est une partie découpée arbitrairement dans le tout. Cette partie est donc souvent désignée en décrivant sa forme et sa position, avec un nom de localisation interne par exemple (*le haut de l'armoire, la pointe du couteau, le sud-ouest de la France*). La partie est toutefois contrainte géométriquement par le fait qu'elle doit être connexe, ce qui n'est pas requis dans les autres méronomies.

Une analyse plus poussée de l'expression des relations de partie à tout en français et en basque, ainsi qu'une formalisation complète de leurs propriétés inférentielles —notamment de la transitivité qui ne s'applique que pour certaines combinaisons— peut être trouvée dans [9]. Le prédicat Part regroupe les six méronomies de façon indifférenciée :

D36  Part(x,y) $\equiv_{def}$ Member(x,y) $\vee$ Subcoll(x,y) $\vee$ Portion(x,y) $\vee$ Subst-Wh(x,y) $\vee$ Component(x,y) $\vee$ Piece(x,y)         "x est une partie de y"

## 3.2 Orientation

La formalisation des processus orientationnels s'appuie sur les outils mis en place au niveau géométrique pour manipuler les concepts d'orientation et prend également en considération des propriétés fonctionnelles directement liées aux entités. Nous présentons dans la suite les définitions formelles proposées pour rendre compte des orientations intrinsèques verticale et frontale des entités. Nous montrons ensuite la manière dont cette modélisation des concepts d'orientation intervient dans la spécification du contenu sémantique des prépositions spatiales externes *devant/derrière*.

### 3.2.1 Les orientations intrinsèques

Il nous faut tout d'abord mettre en évidence le fait que, dans de nombreux cas, associer une orientation intrinsèque à une entité revient à dire que, pour des raisons fonctionnelles, une portion particulière de cette entité constitue une extrémité dans la direction considérée (par exemple le goulot d'une bouteille délimite cette bouteille vers le haut).

En nous basant sur cette remarque, nous introduisons une nouvelle fonction partielle mettant en correspondance une extrémité y d'une entité x (et un extrémité z d'une portion de x) avec la direction correspondante D (la relation Exts utilisée dans cet axiome a été définie au niveau géométrique) :

A44  dir-ext(y,z,x)=D $\leftrightarrow$ (Part(y,x) $\wedge$ Part(z,x) $\wedge$ Exts(sref(y),sref(z),sref(x),D))

Dans la suite nous dirons qu'une telle direction est générée par les extrémités y et z de x. Une direction donnée peut être considérée comme constituant la direction intrinsèque supérieure d'une entité si, dans une situation canonique, cette direction coïncide avec la direction supérieure induite par la gravité :

D37  Orient-haut(D,x) $\equiv_{def}$ $\exists$y,z  (dir-ext(y,z,x)=D $\wedge$ Can-Use(x) $\wedge$ (In-Use(x) > dir-ext(y,z,x)=haut-grav))

Dans cette définition le prédicat "Can-Use" indique que l'entité x a un usage canonique. Le prédicat "In-Use" associé à un mécanisme d'implication non-monotone (> dénotant une implicature) nous permet de restreindre la coïncidence entre directions aux situations dans lesquelles l'entité x donne lieu à une utilisation canonique.

Une formule similaire caractérise ce qu'est une orientation intrinsèque inférieure, la relation entre cette notion et celle d'orientation supérieure préalablement introduite étant également spécifiée :

D38  Orient-bas(D,x) $\equiv_{def}$ $\exists$y,z  (dir-ext(y,z,x)=D $\wedge$ Can-Use(x) $\wedge$ (In-Use(x) > dir-ext(y,z,x)=bas-grav))

D39  bas-grav $=_{def}$ - (haut-grav)

Le fonctionnement de l'orientation frontale fait appel à des mécanismes plus complexes. En fait, nous distinguons trois cas d'orientations frontales intrinsèques qui ne sont cependant pas mutuellement exclusifs.

Le premier cas (êtres humains, animaux, flèches, voitures, véhicules en général...) couvre les situations dans lesquelles l'orientation frontale d'une entité x découle de ce que Vandeloise appelle l'"orientation générale" de x [32] et qui dépend de plusieurs facteurs, parmi lesquels, la direction frontale, la direction du déplacement et la disposition des organes perceptifs:

D40  Orient-avant1(D,x) $\equiv_{def}$ $\exists$y,z dir-ext(y,z,x)=D $\wedge$ Orient-gen(x,D)

Un second type d'orientation frontale (qualifié d'orientation en tandem) regroupe l'ensemble des entités dont la direction frontale coïncide, lors d'une utilisation canonique, avec la direction frontale de l'utilisateur (chaises, voitures, vêtements...). A travers cette seconde règle, nous indiquons donc qu'une direction spécifique d'une entité x constitue une direction frontale de type 2 si la direction frontale d'une entité quelconque utilisant x d'une manière canonique coïncide avec cette direction de x :

D41  Orient-avant2(D,x) $\equiv_{def}$ $\exists$y,z  (dir-ext(y,z,x)=D  $\wedge$ Can-Use(x)  $\wedge$ $\forall$u,D' ((Utilise(x,u) $\wedge$ Orient-avant1(D',u)) > D'=dir-ext(y,z,x)))

La troisième et dernière règle caractérise les entités dont la direction frontale est opposée, lors d'un usage canonique, à la direction frontale de l'utilisateur (armoires, ordinateurs, télévisions...) :

D42  Orient-avant3(D,x) $\equiv_{def}$ $\exists$y,z  (dir-ext(y,z,x)=D  $\wedge$ Can-Use(x)  $\wedge$ $\forall$u,D' ((Utilise(x,u) $\wedge$ Orient-avant1(D',u)) > D'=-dir-ext(y,z,x)))

Enfin, nous exprimons au moyen des formules ci-dessous, que toute entité possédant une orientation frontale intrinsèque obéit à l'un des trois cas de figure distingués ci-dessus et que les directions avant et arrière constituent des directions opposées :

D43  Orient-avant(D,x) $\equiv_{def}$  Orient-avant1(D,x)  $\vee$ Orient-avant2(D,x)  $\vee$ Orient-avant3(D,x)

A45  Orient-avant(D,x) $\leftrightarrow$ Orient-arriere(-D,x)

La formalisation de l'orientation intrinsèque latérale dont nous ne donnerons pas ici le détail fait appel à des représentations plus complexes que celles introduites pour modéliser l'orientation frontale (ces dernières étant elles-mêmes plus complexes que celles associées à l'orientation verticale). Cette propriété de nos outils formels reflète bien les observations effectuées par les psycholinguistes à propos de l'acquisition et de la manipulation des notions d'orientation [26].

### 3.2.2 Orientation et sémantique des prépositions spatiales externes

Nous illustrons, dans la suite, la manière dont les outils formels élaborés pour rendre compte des processus orientationnels peuvent être mis en œuvre pour représenter le contenu sémantique de certains marqueurs spatiaux. Pour cela nous examinons les définitions formelles associées aux prépositions spatiales externes *devant/derrière*.

Une entité y est décrite comme se trouvant située (intrinsèquement) devant une entité x si y est incluse dans la portion d'espace située devant x (c'est-à-dire dans la portion d'espace délimitée au moyen de x et de sa direction frontale intrinsèque). Afin de saisir une telle notion, nous introduisons le prédicat In-sp(y,x,D) qui indique qu'une entité y est incluse dans l'espace délimité au moyen de l'entité x et de la direction D. D'un point de vue formel, ceci est exprimé en posant qu'une relation m$_i$ ou > existe entre les référents spatiaux de y et de x dans la direction D[8] :

---

[8]Cette spécification de "In-sp" est suffisante pour des entités parallélépipédiques, sphériques et cylindriques. La prise en compte d'entités ayant des formes plus compliquées (telles que des amphithéâtres, des arches) accroîtrait la complexité de la formalisation. Cette dernière possibilité a été testée sur un

D44  In-sp(y,x,D) ≡$_{def}$ m$_i$ >(sref(y),sref(x),D)

Nous pouvons maintenant caractériser le fait qu'une entité y est située intrinsèquement devant une entité x en indiquant que y se trouve dans l'espace délimité au moyen de x et de la direction D et que, de plus, cette dernière direction constitue la direction frontale intrinsèque de x :

D45  Etre-devant-i(y,x,D) ≡$_{def}$ Orient-avant(D,x) ∧ In-sp(y,x,D)

L'usage déictique de la préposition *devant* diffère de son usage intrinsèque par le fait que la direction sous-jacente est induite par le locuteur décrivant la scène située devant lui et non par le site lui-même :

D46  Etre-devant-d(y,x,D) ≡$_{def}$ ∃s (Orient-avant(-D,s) ∧ s≠x ∧ s≠y ∧ Speaker(s) ∧ In-sp(y,x,D) ∧ Etre-devant-i(x,s,-D))

Le fait que le locuteur soit placé devant le site auquel il donne une orientation frontale signifie que nous considérons ici une configuration en miroir (entre le locuteur orienteur et le site). Ceci est exprimé par le signe négatif associé à la direction apparaissant dans le prédicat "Orient-avant". En fait, les interactions de type miroir sont très fréquentes en français par opposition aux orientations en tandem qui semblent moins souvent utilisées.

Les définitions formelles associées à la préposition *derrière* sont très similaires à celles proposées pour *devant*, les principales différences concernant la nature des orientations sous-jacentes. Précisons que cette modélisation des concepts d'orientation a également permis de rendre compte de la sémantique d'un certain nombre de lexèmes utilisés pour désigner les diverses portions d'une entité et appelés noms de localisation interne (ex : *haut, bas, avant, arrière, dessus, dessous, devant, derrière*...). On trouvera une formalisation de la sémantique de ces éléments lexicaux dans [4] et dans [5].

En nous focalisant sur les emplois statiques des prépositions orientationnelles, nous avons délibérément laissé de côté un élément du contexte susceptible de jouer un rôle majeur dans l'interprétation de ces prépositions à savoir le déplacement. Il semble donc important de noter que, parallèlement à cette description formelle de l'orientation statique dans la langue, une analyse des interprétations dynamiques des prépositions orientationnelles a été entreprise [7, 22]. Elle devrait, à terme, permettre d'aboutir à la définition d'un cadre théorique unifié pour la représentation des notions d'orientation dans la langue.

### 3.3 Support et préposition *sur*

La position des entités sur l'axe vertical constitue un critère essentiel au moment de différencier les diverses configurations spatiales auxquelles permet de se référer la préposition *sur*. Si la cible se trouve située plus haut que le site (a) nous parlerons de sur1. La situation dans laquelle le site est placé au même niveau que le site (b) sera désignée par sur2. Enfin sur3 s'applique lorsque la cible est située plus bas que le site (c).

  *(a) Le livre est sur la table*
  *(b) L'affiche est sur le mur*
  *(c) La mouche est sur le plafond*

Du point de vue géométrique, ces configurations spatiales donnent lieu à trois types de contact entre individus (notés respectivement Cont1, Cont2, et Cont3). Ainsi, Cont1 correspond aux situations dans lesquelles une zone z1 de la surface de x est en contact avec une zone z2 de la surface de y, z1 étant située plus haut que z2 (le prédicat Zonecont(z1,x,y) caractérise la zone de contact z1 entre x et y, c'est-à-dire la portion maximale de l'enveloppe de x en contact faible avec y) :

D47  Cont1(x,y) ≡$_{def}$ Cont(x,y) ∧ ∃z1,z2 (Zonecont(z1,x,y) ∧ Zonecont(z2,y,x) ∧ Plus_haut(z1,z2))

---

certain nombre d'entités. On a pu en particulier montrer que plusieurs propriétés inférentielles intéressantes caractérisant la version initiale du prédicat "In-sp" n'étaient plus vérifiées.

La comparaison des positions relatives des deux zones de contact entre les référents spatiaux des entités concernées (et non les positions relatives des référents spatiaux des entités elles-mêmes, afin de traiter correctement le cas d'une personne assise sur une chaise) permet donc de classer dans l'un des trois cas mentionnés plus haut les diverses configurations décrites par la préposition *sur*.

Hormis ces caractéristiques géométriques (positions relatives des zones de contact), la sémantique de la préposition *sur* fait également appel à deux concepts fonctionnels importants, à savoir la notion de "catégories comparables" et celle de "stabilisation".

Deux entités x et y appartiennent à des catégories comparables si elle présentent des dimensions similaires, ce que l'on matérialise par la relation Catcomp(x,y). Cette propriété est calculée en comparant l'extension de x et de y selon les divers axes ou dimensions associés à ces entités. Selon la configuration considérée, l'extension relative des entités dans une dimension particulière peut avoir plus d'importance que leur extension dans les autres dimensions. Par exemple dans le cas d'un sur1 (d) les tailles respectives de la cible et du site selon l'axe vertical sont assez peu contraintes alors que pour des usages de type sur3 il est beaucoup plus difficile d'admettre une extension de la cible selon cette dimension (e). En conséquence, nous faisons appel à trois prédicats Catcomp différents (Catcomp1, Catcomp2, Catcomp3) correspondant aux trois configurations de *sur*. Une spécification complète de la notion de catégories comparables doit enfin tenir compte d'un certain nombre de propriétés liées à la nature et à la fonction des entités.

> *(d) Le vase est sur la nappe*
> *(e) \* Le lustre est sur le plafond*

Le support ou stabilisation constitue un autre concept fonctionnel jouant un rôle important dans la sémantique de la préposition *sur*. Dans notre système, le prédicat Stabilise(x,y) indique qu'une entité x stabilise une entité y et le postulat suivant établit que, contrairement à la relation de contact, la stabilisation est transitive :

A46  (Stabilise(x,y) $\wedge$ Stabilise(y,z)) $\rightarrow$ Stabilise(x,z)

Une entité stable par nature (ex : le sol) est qualifiée de stabilisateur intrinsèque, toute entité n'appartenant pas à cette catégorie devant être stabilisée par une autre entité en contact avec elle :

A47  $\neg$Stabilisateur_Intrinseque(x) $\rightarrow$ $\exists$y ($\neg$(y=x) $\wedge$ Stabilise(y,x) $\wedge$ Cont(sref(y), sref(x)))

Un axiome doit être également introduit afin de rendre compte de l'interaction entre relations de partie à tout et processus de stabilisation. Si une partie z d'une entité y stabilise une entité x, alors y stabilise x :

A48  (Part(z,y) $\wedge$ $\neg$Part(x,y) $\wedge$ Stabilise(z,x)) $\rightarrow$ Stabilise(y,x)

Le concept de stabilisation totale est défini en posant qu'une entité y stabilise totalement une entité x si, non seulement y stabilise x mais si, de plus, toute entité z disjointe de y stabilisant directement x est elle-même totalement stabilisée par y :

D48  Stab_tot(y,x) $\equiv_{def}$ Stabilise(y,x) $\wedge$ $\forall$z ((Cont(sref(z),sref(x)) $\wedge$ Stabilise(z,x) $\wedge$ $\neg$O(sref(z),sref(y)) $\rightarrow$ Stab_tot(y,z))

L'ensemble de ces outils géométriques et fonctionnels nous permettent d'introduire la définition suivante pour les configurations de type sur1 :

D49  Sur1(x,y) $\equiv_{def}$ Catcomp1(x,y) $\wedge$ Cont1(sref(x),sref(y)) $\wedge$ Stabilise(y,x)

Cette définition stipule que si y et x appartiennent à des catégories comparables, si la zone de contact de x (avec y) est située plus haut que la zone de contact de y (avec x), et si, de plus, y stabilise x, nous pouvons déduire que *x est sur y*.

Les configurations de type sur2 dans lesquelles la cible est située au même niveau que le site, sont caractérisées au moyen d'une définition similaire, les principales différences concernant le type de contact (Cont2), les catégories comparables (Catcomp2) et la nature du support qui, dans ce cas, doit être total :

D50  Sur2(x,y) $\equiv_{def}$ Catcomp2(x,y) $\wedge$ Cont2(sref(x),sref(y)) $\wedge$ Stab_tot(y,x)

La notion de stabilisation totale introduite ici permet, par exemple, de distinguer la situation dans laquelle une télévision est posée sur une étagère elle-même fixée à un mur (*la télévision est sur le mur*) de celle dans laquelle cette télévision repose sur une table placée contre le mur (#*la télévision est sur le mur*).

La définition associée au cas sur3 diffère de celle associée à sur2 par le type de contact entre les entités concernées ainsi que par le prédicat Catcomp3 relatif aux catégories comparables (ce prédicat est le plus restrictif parmi les divers prédicats "Catcomp", en particulier pour ce qui concerne l'axe vertical) :

D51  Sur3(x,y) $\equiv_{def}$ Catcomp3(x,y) $\wedge$ Cont3(sref(x),sref(y)) $\wedge$ Stab_tot(y,x)

### 3.4 Contenance et préposition *dans*

En ce qui concerne *dans*, notons tout d'abord que la relation géométrique d'inclusion qui est souvent considérée comme la formalisant, ne relie en général pas entre eux les référents spatiaux des entités. Lorsque *le livre est dans l'armoire*, le livre et l'armoire ne partagent aucune portion de matière, leurs référents spatiaux ne se recouvrent donc pas. Le référent spatial du livre est inclus dans le référent spatial du morceau d'espace qu'est l'*intérieur* de l'armoire. On peut noter que la fonction géométrique de fermeture convexe, fréquemment utilisée aussi, ne définit qu'imparfaitement l'intérieur car une concavité quelconque ne correspond pas forcément à un intérieur, comme cela est visible sur la figure 1. Dans le cas d'objets dont la fonction est de contenir (*vase*, *boîte*), une concavité doit être elle-même "contenante" pour constituer un intérieur [19].

La propriété de contenance peut être décrite comme la restriction du mouvement potentiel du contenu [32]. Elle se fonde en particulier sur l'opposition à la gravité mais se distingue de la notion de support par des restrictions supplémentaires concernant les mouvements latéraux (d'où la différence entre les expressions *sur un tabouret* et *dans un fauteuil*). La notion de contenance s'avère donc particulièrement importante pour la sémantique de *dans*, même si l'expression *x est dans y* n'implique pas obligatoirement que y contienne x (*l'oiseau est dans le ciel* ne suppose aucun phénomène de contenance). Nous ne donnons pas ici de formalisation de la notion de contenance, mais nous axiomatisons la notion d'intérieur.

Nous venons de voir que l'intérieur d'une entité contenante correspond à l'ensemble de ses concavités contenantes. Les intérieurs des entités non contenantes sont définis exclusivement par leur forme. Ils sont de trois types : On distingue d'une part le cas des objets éparpillés (collections comme dans *le chien est dans la foule*) ou déterminant un volume sans le remplir ni le délimiter (*l'oiseau est dans l'arbre*). Leurs intérieurs sont alors définis grâce à la fonction de contour, ou "outline", introduite dans [19] (voir aussi [33]). D'autre part, on distingue les cas des objets ou morceaux de matière non solides entourant complètement leur intérieur, qui est souvent temporaire et créé par l'enchâssement de la cible dans le site (*le poisson est dans la mer/l'eau*). Le référent spatial de ces intérieurs est alors une composante connexe du complément du référent spatial de l'entité. Enfin, les intérieurs des lieux sont définis un peu plus arbitrairement par un morceau d'espace limité latéralement par des verticales passant par les frontières du lieu et verticalement par le lieu lui-même et par un plan horizontal situé "suffisamment" haut au-dessus du lieu (ceci étant assez compliqué à représenter d'un point de vue géométrique, cette contrainte n'est pas considérée dans l'axiome A49). Les morceaux d'espace n'ont évidemment pas d'intérieur, ils jouent eux-mêmes directement ce rôle. Nous faisons l'hypothèse que les substances ne définissent pas non plus d'intérieur.

La fonction "int" vérifie les axiomes suivants[9] :

A49  y=int(x) $\rightarrow$ ((Obj(x) $\vee$ Mat(x) $\vee$ Loc(x)) $\wedge$ Sp-port(y) $\wedge$ Depend(y,x) $\wedge$ (t=int(x) $\rightarrow$ y=t) $\wedge$ ICont(sref(x),sref(y)) $\wedge$ ($\neg$Loc(x) $\rightarrow$ (P(i(sref(y)),preint(sref(x))) $\wedge$ (Container(x) $\wedge$ $\Diamond\exists$z Contain(y,z)) $\vee$ sref(y)=outline(sref(x)) $\vee$ Con-Comp(sref(y), -sref(x)))))

A50  (t=int(x) $\wedge$ u=int(y) $\wedge$ Part(x,y) $\wedge$ Rest(y,x,r)) $\rightarrow$ P(sref(t), sref(u)+sref(r)))

A51  (t=int(x) $\wedge$ u=int(y) $\wedge$ P(i(sref(x)), sref(u)) $\rightarrow$ P(sref(t), sref(u)+sref(y)))

---

[9] La fonction géométrique preint est telle que preint(x) dénote la fermeture convexe de l'individu x, moins x ; le prédicat géométrique Con-Comp(x,y) établit que x est une composante connexe de l'individu y ; $\Diamond$ est l'opérateur modal de possibilité ; Rest(y,x,r) établit que r est la partie de y complémentaire de x dans y.

On peut décrire la sémantique de *dans* en distinguant trois types de configurations spatiales. Tous les exemples que nous venons d'évoquer illustrent les deux premiers cas à savoir les situations dans lesquelles le référent spatial de la cible est inclus dans le référent spatial de l'intérieur du site (du site lui-même dans le cas d'un morceau d'espace) ou bien le recouvre. Dans le premier cas —situation prototypique— cette inclusion est totale (*le livre est dans l'armoire*). Nous appellerons ce cas "dans-total". Dans le second cas, il n'y a que recouvrement, ou "inclusion partielle" (*la cuillère est dans la tasse*). Ce cas sera dénommé "dans-partiel".

Le troisième cas semble, à première vue, assez différent : il décrit une relation de partie à tout entre les deux entités et donc le référent spatial de la cible est directement inclus dans le référent spatial du site. *L'escalier est dans la maison* et *l'homme est dans la foule* en constituent des exemples. Nous appellerons ce troisième cas "dans/partie-de". Toute méronomie ne peut toutefois être décrite au moyen de la préposition *dans*. Par exemple, si *le cerveau est dans la tête* est acceptable, \**le nez est dans la tête* ne l'est pas [32]. On peut rendre compte de ce phénomène à travers un principe que nous appelons "de contraste". Une expression spatiale décrivant une méronomie met en évidence la position de la partie par rapport au tout : on considère donc qu'elle relie la partie au tout auquel on aurait ôté, par contraste, cette partie. Le dernier exemple envisagé est inacceptable car le nez n'est inclus dans aucune concavité de "la tête diminuée du nez". Le principe de contraste explique aussi l'emploi de phrases situant le tout dans la partie et qui pourraient donc sembler paradoxales à première vue (*la noix/l'escargot est dans sa coquille*). Toutefois, il n'est requis que lorsque la méronomie est un cas de "composant / assemblage" ou "morceau / tout", et seulement lorsque la cible et le site sont des objets ou des morceaux de matière. Dans les autres cas, toute instance de méronomie peut être décrite par l'usage de la préposition *dans* (*Le Cotentin est dans le département de la Manche, Paul est dans le jury*)

Dans cette première analyse, nous n'avons pas considéré d'assez près la particularité des usages de *dans* où la cible n'est pas un objet ou un morceau de matière, mais un morceau d'espace ou un lieu. Un morceau d'espace situé dans un objet ou un morceau de matière (*il y a un trou dans ce morceau de fromage*) décrit en fait une relation de partie à tout (de type "morceau / tout") entre la cible et l'intérieur du site. On peut à la rigueur situer un morceau d'espace dans un autre (*l'intérieur de la boîte est dans l'intérieur de l'armoire*), mais cela ne décrit qu'une simple inclusion. Un lieu ne peut être situé que dans un autre lieu. S'il ne s'agit pas d'une méronomie, il s'agit alors de la description d'une situation d'enclave (*l'île est dans la mer*, *Saint-Marin est en Italie*), pour laquelle il faut remarquer que le contact entre la cible et le site est nécessaire. Nous considérerons que tous ces cas sont des occurrences de "dans-total".

La définition obtenue pour "dans total" est donc :

D52  TDs(x,y) ≡$_{def}$ [(Obj(x) ∨ Mat(x)) ∧ (Obj(y) ∨ Mat(y) ∨ Loc(y)) ∧ P(i(sref(x)), sref(int(y))))] ∨ [(Obj(x) ∨ Sp-port(x)) ∧ Sp-port(y) ∧ P(i(sref(x)),sref(y))] ∨ [Sp-port(x) ∧ (Obj(y) ∨ Mat(y)) ∧ (Piece(x,int(y)) ∨ x=int(y))] ∨ [Loc(x) ∧ Loc(y) ∧ ∀z (EC(sref(z),c(sref(ground)•(-sref(x)))) → EC(sref(z), sref(y)))]

Pour "dans-partiel", on a :

D53  PDs(x,y) ≡$_{def}$ [(Obj(x) ∨ Mat(x)) ∧ (Obj(y) ∨ Mat(y) ∨ Loc(y)) ∧ O(i(sref(x)), sref(int(y))))] ∨ [(Obj(x) ∨ Mat(x)) ∧ Sp-port(y) ∧ O(i(sref(x)),sref(y))]

Et la définition de "dans/partie-de" est :

D54  DPt(x,y) ≡$_{def}$ [Part(x,y) ∧ ((((Component(x,y) ∨ Piece(x,y)) ∧ (Obj(x) ∨ Mat(x)) ∧ (Obj(y) ∨ Mat(y))) → ∃z (Rest(x,y,z) ∧ TDs(x,z)))] ∨ [(Component(y,x) ∨ Piece(y,x)) ∧ (Obj(x) ∨ Mat(x)) ∧ (Obj(y) ∨ Mat(y)) ∧ ∃z (Rest(y,x,z) ∧ TDs(y,z)))]

## 4 Inférences et pragmatique

Comme cela a déjà été noté, l'analyse proposée ici ne se limite pas à la seule représentation du contenu sémantique mais intègre également la dimension inférentielle des processus interprétatifs. Nous détaillons dans la suite certaines des inférences obtenues dans le cadre de la théorie formelle introduite. Nous montrons ensuite que le résultat de ces inférences n'est pas toujours satisfaisant et qu'une meilleure adéquation au raisonnement humain nécessite la prise en compte de données pragmatiques liées principalement au contexte et à la connaissance du monde.

## 4.1 Inférences

### 4.1.2 Les prépositions spatiales externes

Nous étudions ici des énoncés composés de deux phrases contenant chacune la préposition spatiale externe *devant* et nous examinons les déductions transitives obtenues à partir de leurs représentations formelles. Plusieurs cas de figure doivent être distingués en fonction de l'interprétation déictique ou intrinsèque des relations spatiales en présence. Une description détaillée de ces divers cas de figure (intrinsèque/intrinsèque, déictique/déictique, intrinsèque/déictique) est proposée dans [5]. Nous considérons dans la suite un énoncé combinant deux prépositions *devant* interprétées intrinsèquement :

    *Le tabouret est devant le fauteuil*
    *Le fauteuil est devant Max*

A partir des définitions introduites pour la préposition *devant* il est possible d'associer à cet énoncé la représentation formelle ci-dessous dans laquelle les constantes t, f et m identifient respectivement le tabouret, le fauteuil et Max :

    Etre-devant-i(t,f,d1)
    Etre-devant-i(f,m,d2)

Le prédicat "In-sp" apparaissant dans la définition de "Etre-devant" nous permet de déduire les relations de Allen suivantes entre les référents spatiaux de t, f et m :

    mi>(sref(t),sref(f),d1)
    mi>(sref(f),sref(m),d2).

Il est important de noter que le processus déductif est conditionné par un paramètre fondamental à savoir l'identité des directions d1 et d2 associées aux deux relations "Etre-devant". Si ces directions coïncident (ce qui est formellement exprimé par d1=d2) il est alors possible, sur la base des axiomes associés aux relations d'Allen (nous utilisons ici le théorème $\forall$ x,y,z (mi>(x,y,D) $\wedge$ mi>(y,z,D) $\rightarrow$ >(x,z,D))), de déduire la relation >(sref(t),sref(m),d2) qui, associée à la définition de "In-sp", permet d'inférer In-sp(t,m,d2). En combinant ce fait à la formule Orient-avant(d2,m), contenue dans la définition de Etre-devant-i(f,m,d2) et dénotant l'orientation intrinsèque frontale de m, on obtient finalement :

    Etre-devant-i(t,m,d2) $\leftrightarrow$ Orient-avant(d2,m) $\wedge$ In-sp(t,m,d2)

A partir des deux phrases citées plus haut et de la contrainte additionnelle concernant la coïncidence des directions frontales intrinsèques du fauteuil et de Max nous parvenons donc à établir que *le tabouret est devant Max*.

Indiquons que les cas de figure déictique/déictique et intrinsèque/déictique donnent lieu à des processus inférentiels similaires, la coïncidence des directions associées à chacune des relations spatiales constituant chaque fois une condition indispensable pour l'application de la transitivité.

### 4.1.2 Cas d'une préposition spatiale interne, *dans*

L'étude détaillée des cas de transitivité de *dans* montre combien l'analyse complexe que nous avons proposée, combinant plusieurs sens d'intérieur, trois situations spatiales et plusieurs classes pour la cible et le site était nécessaire. En effet, *dans* est loin d'être transitive dans tous les cas, alors que la transitivité doit être admise si cette préposition est simplement modélisée par l'inclusion. Nous ne présentons ici que quelques cas. L'étude complète, y compris la démonstration des différents théorèmes, peut être trouvée dans [33].

"Dans-total" entre deux objets et un lieu est transitif :
    *Paul est dans la maison*, *La maison est dans l'île* ⊢ *Paul est dans l'île*

Nous allons en esquisser rapidement la démonstration. Puisque Paul et la maison sont des objets, et que l'île est un lieu, l'antécédent est interprété par : TDs(paul,maison) $\wedge$ TDs(maison,île), ce qui donne :

$P(i(sref(paul)), sref(int(maison))) \wedge P(i(sref(maison)), sref(int(île)))$.

La seconde clause donne, par l'axiome A51 :

$P(sref(int(maison)), sref(int(île)) + sref(île))$

En supposant ensuite que $\neg O(sref(int(maison)), sref(île))$, par un postulat général sur la séparation des référents spatiaux des lieux avec ceux des entités d'autre type, et à l'aide du théorème $(P(x,y+z) \wedge \neg O(x,y)) \rightarrow P(ix,z)$ nous obtenons :

$P(i(sref(int(maison))), sref(int(île)))$

Puisque $(P(x,iy) \rightarrow P(ix,iy))$ et $\forall x \forall y (P(ix,y) \rightarrow P(ix,iy))$ sont aussi des théorèmes, par la transitivité de P, nous concluons enfin la formule suivante qui est bien l'interprétation du conséquent :

$P(i(sref(paul)), sref(int(île)))$

On peut montrer que "dans-total" entre trois objets x, y et z est aussi transitif, si l'on admet $\neg O(sref(int(y)), sref(z))$, ce qui est réaliste dans la plupart des contextes.[10]

Mais "dans-total" entre un objet et deux lieux n'est pas transitif, ce qui peut encore une fois être démontré :

*Paul est dans l'île*, *L'île est dans la mer* ⊬ #*Paul est dans la mer*

La non-transitivité est due au fait que les référents spatiaux des intérieurs de deux lieux en relation de "dans-total" ne sont pas inclus l'un dans l'autre, ils ne se recouvrent même pas. Un "dans-total" entre lieux correspond à une sorte de relation d'entourement, qui ne situe donc pas le premier lieu par rapport à l'intérieur du second. On peut noter que s'il s'était agit d'une relation "dans/partie-de" entre lieux (*Paul est dans le Tarn*, *le Tarn est dans Midi-Pyrénées*), la transitivité serait assurée.

On peut également montrer que la combinaison d'un "dans-total" entre une portion d'espace et un objet, et d'un "dans-total" entre deux objets n'est pas valide à cause de la présence d'une relation de partie à tout dans le premier "dans-total" qui ne peut pas être transmise :

*Il y a un trou dans le drap*, *Le drap est dans le tiroir* ⊬ #*Il y a un trou dans le tiroir*

"Dans-partiel" n'est jamais transitif, à cause de la non-transitivité de O. La transitivité de "dans/partie-de" varie selon les cas car elle repose sur la transitivité des méronomies et de "dans-total", or on a pu noter plus haut que ces transitivités ne sont pas toujours valides.

### 4.2 Pragmatique

Diverses lois et conventions pragmatiques agissent sur les représentations et inférences obtenues au niveau sémantique. Au delà des connaissances purement fonctionnelles, elles se basent sur la connaissance du monde (en particulier la connaissance des situations typiques) et sur les informations issues du contexte. Les lois que nous envisageons à ce niveau peuvent être considérées comme des adaptations au domaine de l'espace de lois plus générales (telles que les principes de coopérativité de Grice).

Tout d'abord, les lois pragmatiques peuvent nous conduire à déduire (souvent par "implicature") plus d'informations qu'il n'y en a effectivement dans le texte et donc plus que n'en donnent les deux premiers niveaux du système. Par exemple, la phrase *Marie est dans la voiture* est généralement interprétée comme *Marie est dans l'habitacle*, écartant par là-même la solution alternative décrite par *Marie est dans le coffre*.

Inversement, ces règles peuvent amener le système à écarter certaines expressions (par exemple, des expressions inférées aux niveaux précédents) qui, bien que correctes d'un point de vue purement sémantique, ne sont pas pour autant inférées car contredisant certaines données ou connaissances pragmatiques. Ainsi, si nous savons que *Marie est dans le coffre de la voiture*, alors la phrase *Marie est dans la voiture* n'est pas fausse et cependant on ne l'utilisera (en général) pas pour répondre à la question *où se trouve*

---
[10]La transitivité ne s'applique pas lorsque Part(x,z) est vérifié.

*Marie ?* puisque dans la plupart des contextes on sait qu'elle sera interprétée comme *Marie se trouve dans l'habitacle*.

Un principe dit de "**fixation**" sous-tend le fonctionnement des exemples cités ci-dessus. Ce principe, introduit dans [32], stipule que l'usage typique d'un objet fixe certaines des ses caractéristiques. Par exemple, l'avant et l'arrière d'une voiture sont fixés par la direction usuelle —et non réelle ou effective— de son mouvement. En fait, de nombreux cas d'orientation intrinsèque sont déterminés de cette façon. Toutefois, l'importance de ce principe peut être telle (par exemple dans le cas de l'orientation) que ses conséquences ne sont jamais remises en cause et qu'il est alors justifié de les prendre en compte au niveau fonctionnel. Ceci illustre la complexité des rapports entre sémantique et pragmatique et l'illusion de vouloir tracer une frontière stricte entre ces deux domaines.

De nombreux autres principes rentrent dans la catégorie des principes restrictifs. Le principe de "cible maximale", qui est en fait un cas particulier de la maxime de quantité indique qu'une relation spatiale situe plutôt le tout que la partie. Le principe symétrique de "site minimal" stipule que plus le site est restreint, plus la localisation est précise. L'application de tels principes doit évidemment être limitée pour éviter que l'on aboutisse à des aberrations (on ne localisera pas un plongeur en disant qu'il est dans son scaphandre), et leurs interactions doivent être contrôlées.

Il reste à mentionner ici un troisième type de facteur pragmatique qui conduit au relâchement ou à la suppression de conditions introduites dans les définitions sémantiques. La possibilité de supprimer certaines conditions apparaissant au niveau sémantique dépend largement du principe gricéen de pertinence, qui indique que si une relation est plus pertinente qu'une autre, la première ne peut être neutralisée au profit de la seconde. C'est en fait cet important phénomène qui contrôle l'acceptabilité de l'imprécision d'une relation spatiale suivant les contextes.

Il est clair que cette partie de notre système doit faire l'objet d'une analyse plus détaillée qui mette en évidence l'ensemble des principes pragmatiques nécessaires au traitement des expressions spatiales et surtout l'articulation entre ces principes. Il ne nous est donc pas possible ici d'en proposer une formalisation.

## Conclusion

L'analyse systématique du contenu sémantique et du fonctionnement des marqueurs spatiaux a permis de mettre en évidence les nombreuses propriétés de l'espace linguistique. Les structures conceptuelles qui sous-tendent cet espace linguistique font appel à une géométrie dont les caractéristiques sont bien souvent en opposition avec les fondements et les principes mêmes de la géométrie cartésienne : localisation relationnelle, imprécision/incomplétude, granularité variable... On a vu, par ailleurs, que les seules données géométriques ne suffisaient pas à saisir le contenu sémantique des marqueurs spatiaux et qu'il était nécessaire de prendre en compte diverses notions liées à la fonction des entités ou bien à la pragmatique. Ces observations nous ont donc amenés à poser les bases d'une véritable géométrie cognitive ainsi qu'à élaborer une théorie en trois niveaux (respectivement géométrique, fonctionnel et pragmatique) permettant de représenter la signification des expressions spatiales et de produire diverses déductions. L'adéquation entre les résultats de ces déductions et ceux des raisonnements humains valide, dans une certaine mesure, le cadre théorique proposé pour représenter les concepts spatiaux dans la langue.

On peut noter que les éléments théoriques proposés dans cette étude ont aussi contribué à modéliser l'apport de la sémantique lexicale de l'espace dans l'analyse de la structure de discours, en particulier pour des textes qui décrivent des trajectoires [2].

Indiquons enfin que plusieurs expérimentations sont développées en collaboration avec des psycholinguistes, dans un but de validation de la partie sémantique de ce travail et d'enrichissement de son versant pragmatique. Ces expérimentations visent notamment à vérifier si la complexité des définitions formelles proposées pour les expressions spatiales étudiées est corrélée à la complexité de leur traitement traduite en termes de temps de réponse. Les premiers résultats confirment l'importance des propriétés fonctionnelles des entités dans le fonctionnement des marqueurs orientationnels, et ceci dans des configurations canoniques aussi bien que non canoniques [13].

# Bibliographie